\documentclass[a4paper,conference]{IEEEtran}

    \usepackage[noadjust]{cite}
    \usepackage{url}

    \usepackage{amsfonts}
    \usepackage{amsmath}
    \DeclareMathOperator*{\argmin}{arg\,min}

    \usepackage[pdftex]{graphicx}
    \usepackage[caption=false,font=footnotesize]{subfig}
    \usepackage{capt-of}
    \usepackage{array}
    \newcolumntype{P}[1]{>{\centering\arraybackslash}p{#1}}
    \usepackage{float}
    \usepackage{flushend}

\newcommand{\defineAcronym}[1]{%
    \newcounter{#1}%
}

\newcommand{\isAcronymDefined}[3]{%
    \ifcsname c@#1\endcsname{%
        #2%
    }\else{%
        #3%
    }\fi%
}

\newcommand{\isAcronymDefinedAuto}[3]{%
    \isAcronymDefined{#1}{#2}{\defineAcronym{#1}{#3}}%
}

\newcommand{\declareAcronym}[2]{
    \expandafter\newcommand\csname #1\endcsname{%
        \isAcronymDefinedAuto{#1}{#1}{#2 (#1)}}
    \expandafter\newcommand\csname #1s\endcsname{%
        \isAcronymDefinedAuto{#1}{#1s}{#2s (#1s)}}
    \expandafter\newcommand\csname #1x\endcsname{%
        \isAcronymDefinedAuto{#1}{#1}{#2}}
    \expandafter\newcommand\csname #1xs\endcsname{%
        \isAcronymDefinedAuto{#1}{#1s}{#2s}}
}

\newcommand{\declareSuperAcronym}[3]{%
    \expandafter\newcommand\csname #1#2\endcsname{%
        \isAcronymDefinedAuto{#1#2}{%
            #1#2%
        }{%
            {#3~}\isAcronymDefined{#2}{#2}{\csname #2x\endcsname}{~(#1#2)}%
        }%
    }%
    \expandafter\newcommand\csname #1#2s\endcsname{%
        \isAcronymDefinedAuto{#1#2}{%
            #1#2s%
        }{%
            {#3~}\isAcronymDefined{#2}{#2s}{\csname #2xs\endcsname}{~(#1#2s)}%
        }%
    }%
    \expandafter\newcommand\csname #1#2x\endcsname{%
        \isAcronymDefinedAuto{#1#2}{%
            #1#2%
        }{%
            {#3~}\isAcronymDefined{#2}{#2}{\csname #2x\endcsname}%
        }%
    }%
    \expandafter\newcommand\csname #1#2xs\endcsname{%
        \isAcronymDefinedAuto{#1#2}{%
            #1#2s%
        }{%
            {#3~}\isAcronymDefined{#2}{#2s}{\csname #2xs\endcsname}%
        }%
    }%
}

    \usepackage{xcolor}
\usepackage{xparse}

\newcommand{\lucasFont}[1]{%
    \textcolor[rgb]{0.0, 0.0, 0.5}{\textbf{#1}}%
}

\newcommand{\pierreFont}[1]{%
    \textcolor[rgb]{0.5, 0.0, 0.0}{\textbf{#1}}%
}

\newcommand{\francoisFont}[1]{%
    \textcolor[rgb]{0.0, 0.4, 0.0}{\textbf{#1}}%
}

\newcommand{\cedricFont}[1]{%
    \textcolor[rgb]{0.8, 0.8, 0.0}{\textbf{#1}}%
}

\NewDocumentCommand{\comment}{omm}{%
    \IfNoValueTF{#1}{%
        #2%
    }{%
        \underline{#1} $\rightarrow$ #2%
    }%
}

\NewDocumentCommand{\lucas}{om}{%
    \lucasFont{\comment[#1]{#2}}%
}

\NewDocumentCommand{\pierre}{om}{%
    \pierreFont{\comment[#1]{#2}}%
}

\NewDocumentCommand{\francois}{om}{%
    \francoisFont{\comment[#1]{#2}}%
}

\NewDocumentCommand{\cedric}{om}{%
    \cedricFont{\comment[#1]{#2}}%
}

\newcommand{\sizemod}[2][0]{
    \ifnum#1=0
        #2
    \else\ifnum#1=1
        \big#2
    \else\ifnum#1=2
        \Big#2
    \else\ifnum#1=3
        \bigg#2
    \else\ifnum#1=4
        \Bigg#2
    \else
        \PackageError{}{invalid parameter}{}
    \fi\fi\fi\fi\fi
}

\newcommand{\parentheses}[2][0]{
    \sizemod[#1](#2\sizemod[#1])
}

\newcommand{\norm}[2][0]{
    \sizemod[#1]\lVert#2\sizemod[#1]\rVert
}

\newcommand{\expectation}[3][0]{
    \underset{#2}{\mathbb{E}} \sizemod[#1][ #3 \sizemod[#1]]
}

    \declareAcronym{HPE}{Human Pose Estimation}
    
    \declareAcronym{GAN}{Generative Adversarial Network}
    \declareSuperAcronym{C}{GAN}{Conditional}
    \declareSuperAcronym{W}{GAN}{Wasserstein}
    \declareSuperAcronym{DC}{GAN}{Deep Convolutional}
    
    \declareAcronym{NN}{Neural Network}
    \declareSuperAcronym{C}{NN}{Convolutional}
    \declareSuperAcronym{D}{NN}{Deep}
    \declareSuperAcronym{R}{NN}{Recurrent}
    
    \declareAcronym{VAE}{Variational Auto-Encoder}
    \declareAcronym{CVAE}{Conditional Variational Autoencoder}
    
    \declareAcronym{MSE}{Mean Squared Error}
    \declareAcronym{MPJPE}{\emph{mean per joint position error}}
    \declareAcronym{MPJVE}{\emph{mean per joint velocity error}}
    
    \declareAcronym{ERD}{Encoder-Recurrent-Decoder}

    \declareAcronym{ReLU}{Rectified Linear Unit}
    \declareSuperAcronym{L}{ReLU}{Leaky}

    \declareAcronym{PCKh}{Percentage of Correct Keypoints normalized with Head size}
    \declareAcronym{AUC}{Area Under the Curve}

\begin{document}
    
    \title{
        JUMPS: Joints Upsampling Method for Pose Sequences
    }
    \author{
        \IEEEauthorblockN{Lucas Mourot, Fran\c{c}ois Le Clerc, C\'edric Th\'ebault and Pierre Hellier}
        \IEEEauthorblockA{\{Lucas.Mourot, Francois.LeClerc, Cedric.Thebault, Pierre.Hellier\}@InterDigital.com}
        \IEEEauthorblockA{InterDigital, Rennes, France}
    }

    \twocolumn[
        {%
        \renewcommand\twocolumn[1][]{#1}%
        \maketitle
        \begin{center}
            \centering
            \includegraphics[width=\linewidth]{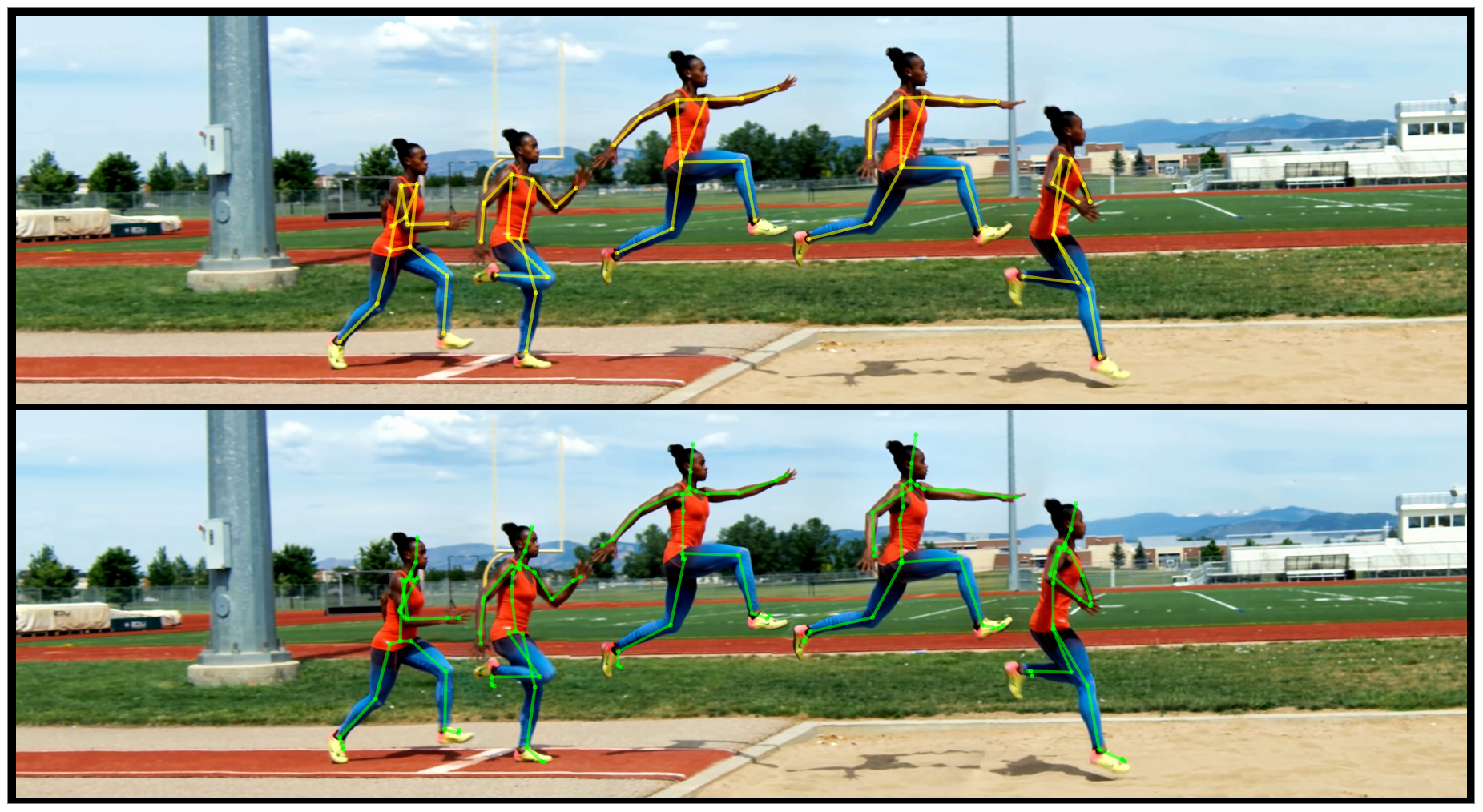}
            \captionof{figure}{%
                Illustration of our joints upsampling method (JUMPS) applied to the task of Human Pose Estimation (HPE): the yellow $12$-joint skeletons (top image) depicts poses estimated by \emph{AlphaPose} \cite{paper_pose_flow} at different timestamps of a long jump. From this set of poses, we compute the green $28$-joint skeletons (bottom image). Our approach upsamples the number of joints in order to enrich the representation, in a spatially and temporally consistent way. In both images, the background has been stitched from a video with a smaller field of view and five images of the jumper at different frames have been overlaid as foreground.
            }
        \end{center}
        }
    ]   
    \thispagestyle{plain}
    \pagestyle{plain}
 
    \begin{abstract}
        Human Pose Estimation is a low-level task useful for surveillance, human action recognition, and scene understanding at large. It also offers promising perspectives for the animation of synthetic characters. For all these applications, and especially the latter, estimating the positions of many joints is desirable for improved performance and realism. To this purpose, we propose a novel method called JUMPS for increasing the number of joints in 2D pose estimates and recovering occluded or missing joints. We believe this is the first attempt to address the issue. We build on a deep generative model that combines a \GAN{} and an encoder. The \GAN{} learns the distribution of high-resolution human pose sequences, the encoder maps the input low-resolution sequences to its latent space. Inpainting is obtained by computing the latent representation whose decoding by the \GAN{} generator optimally matches the joints locations at the input. Post-processing a 2D pose sequence using our method provides a richer representation of the character motion. We show experimentally that the localization accuracy of the additional joints is on average on par with the original pose estimates.
    \end{abstract}
    
    \section{Introduction} \label{section_introduction}
        \HPE{} refers to the problem of predicting joints position (either in $2D$ or $3D$) of a person in an image or video. It has been a topic of active research for several decades, and all state-of-the-art solutions rely on deep learning \cite{paper_openpose, paper_CFA, paper_darkpose, paper_hrnet, paper_unipose}. Even then, the best approaches extract skeletons with a limited number of joints, usually from $12$ to $16$, which is too rough for the movie industry or video games applications. This issue concerns both the $2D$ and $3D$ cases since $3D$ joint extraction almost always relies on $2D$ pose estimation. Moreover, these approaches still fail in the presence of strong foreshortening, left-right ambiguities, (self)-occlusions, or on previously unseen complex poses.\par

In this paper we improve on state-of-the-art \HPE{} solutions by upsampling human joints and inpainting occluded ones, thereby paving the way for downstream applications that require higher skeleton resolution. Starting with a temporal sequence of partially occluded poses, we recover missing joint locations and improve the resolution of the skeleton animation by estimating the positions of additional joints. To the best of our knowledge, no work has been previously proposed to recover missing joints or increase joints resolution of animated skeletons in 2D. We believe that enriching the representation helps in many cases, especially for extremities such as feet/toes and hands. A better extraction of the former provides a better visualisation and understanding of the motion. For instance, extracting the toe in addition to the ankle provides a better sense of feet contacts.\par

To this purpose, we draw inspiration from past research on human pose estimation, human motion modeling and image inpainting based on deep generative models: we leverage a deep generative network that provides an effective prior on spatio-temporal biomechanics. Our model builds on a \GAN{}, which we complement by a front-end encoder to learn a mapping from the human pose space to the \GAN{} latent space. The encoder helps selecting better samples in latent space and stabilizes the training of the \GAN{}.\par

In summary, our paper proposes the following contributions:
\begin{itemize}
    \item a novel method based on deep generative modeling for inpainting $2D$ pose sequences and enriching the joints representation. The method relies on temporal sequence analysis since motion is key to recover missing joints.
    \item a hybrid GAN/autoencoder architecture; we show that the autoencoder is crucial for a better convergence and accuracy.
    \item We show that optimization in latent space is greatly improved by adding a Procrustes alignment at each iteration.
    \item We provide qualitative and quantitative assessments of the effectiveness of our method on the \emph{MPI-INF-3DHP} human pose dataset.
\end{itemize}
 
    \begin{figure}[t]
    \centering
    \includegraphics[width=\linewidth]{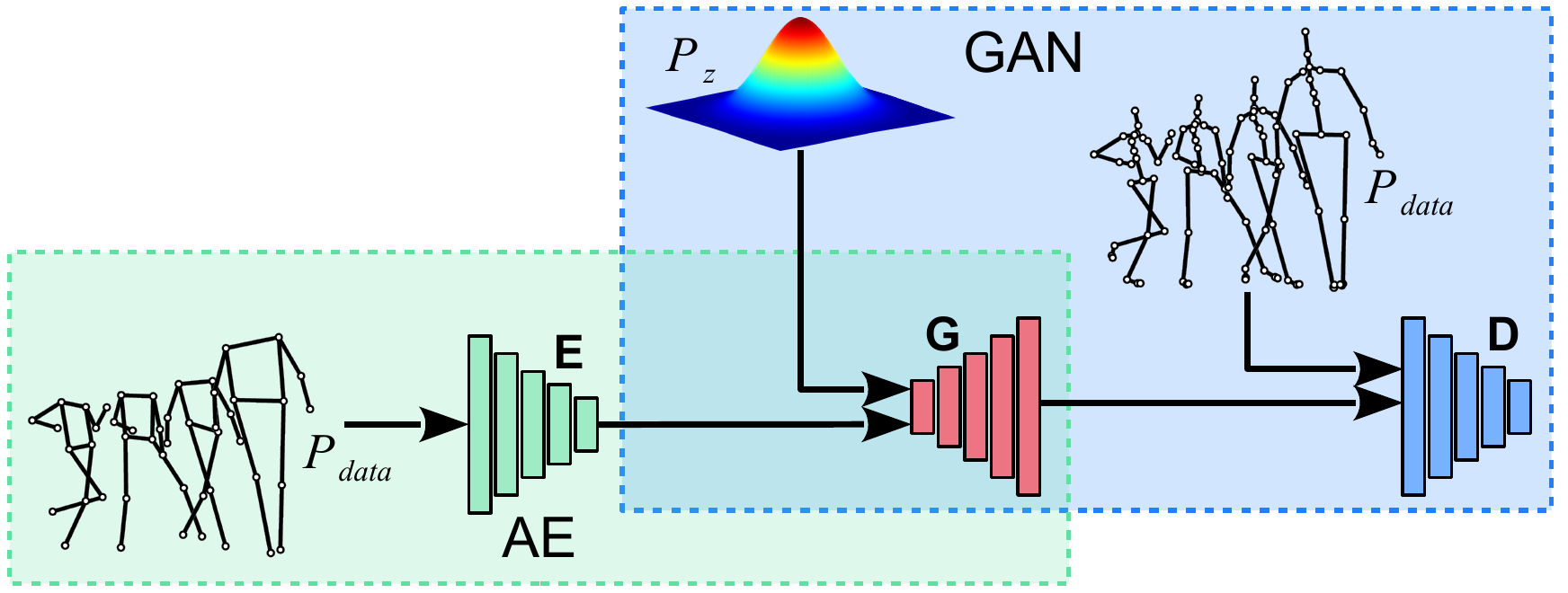}
    \caption{
        Coarse representation of our deep generative network; the upper right part (blue) is the basis of our model: a Generative Adversarial Network (GAN) with generator $G$ and discriminator $D$; the bottom left part (green) depicts how the encoder $E$ and the generator $G$ together yield an autoencoder (AE) scheme; $P_{z}$ and $P_{data}$ denote the prior and data distribution respectively.
    }
    \vspace{-0.1in}
    \label{figure_network}
\end{figure}
    \begin{figure}[t]
	\centering
	\includegraphics[width=3.07in]{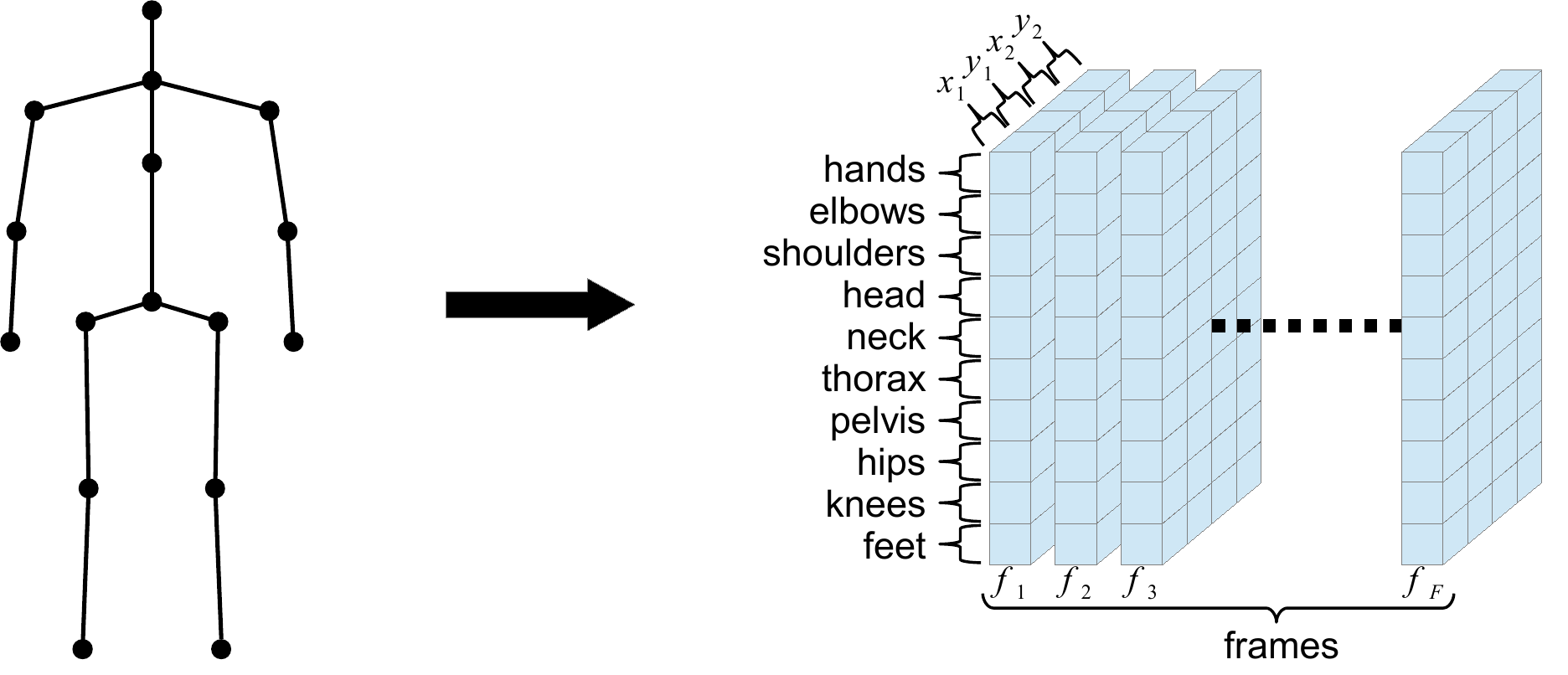}			
	\caption{Illustration of how our subnetworks internally represent a pose sequence whose topology is depicted on the left part. The right part shows how joints coordinates are arranged, with body parts ordered following the human skeleton from hands to feet.}
	\vspace{-0.16in}
	\label{figure_data_representation}
\end{figure}

    \begin{figure*}[t]
    \centering
    
    \subfloat[Encoder]%
    {\includegraphics[width=0.33\textwidth]%
    {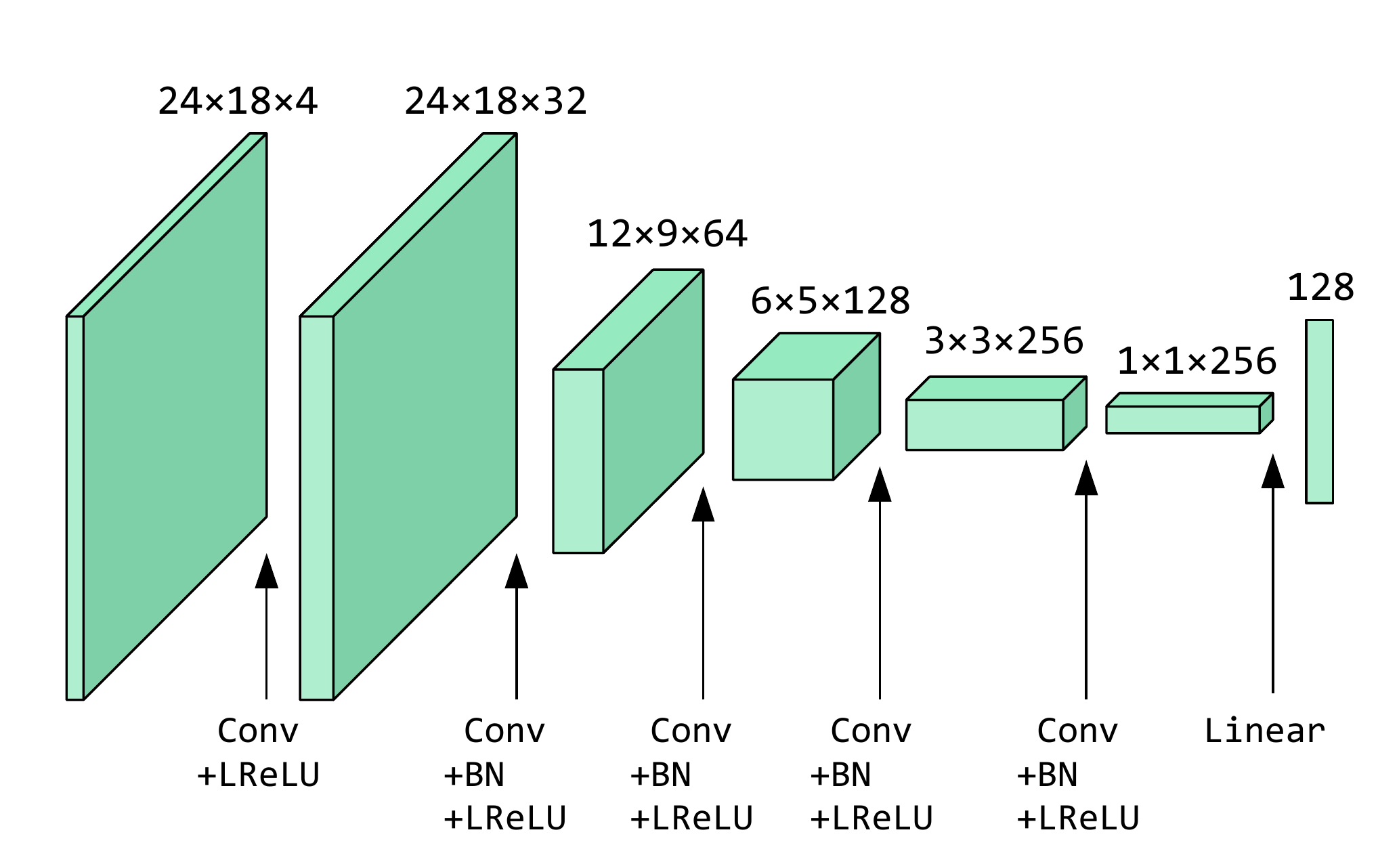}%
    \label{figure_architecture_encoder}}
    \hfil
    \subfloat[Generator]%
    {\includegraphics[width=0.33\textwidth]%
    {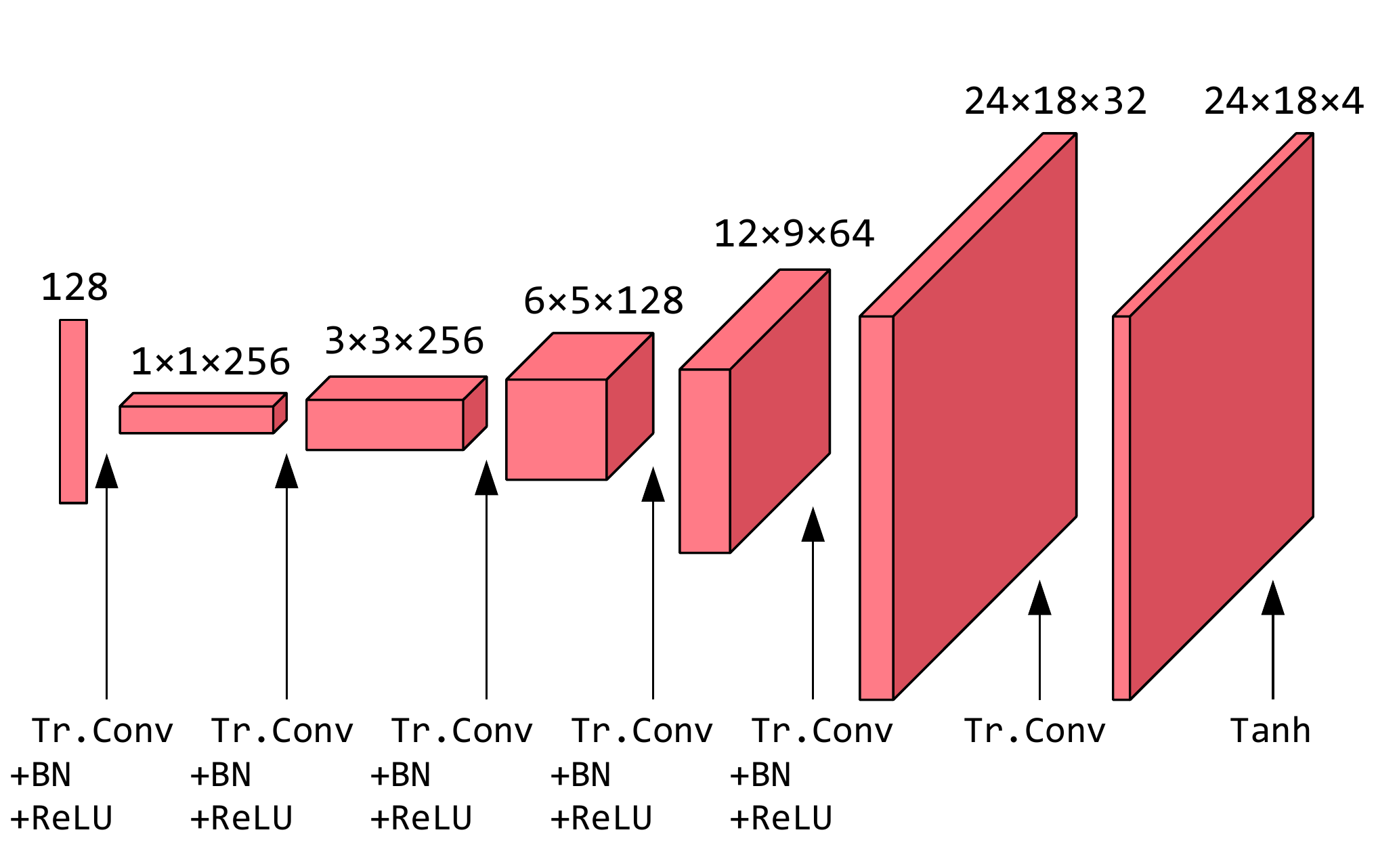}%
    \label{figure_architecture_generator}}
    \hfil
    \subfloat[Discriminator]%
    {\includegraphics[width=0.33\textwidth]%
    {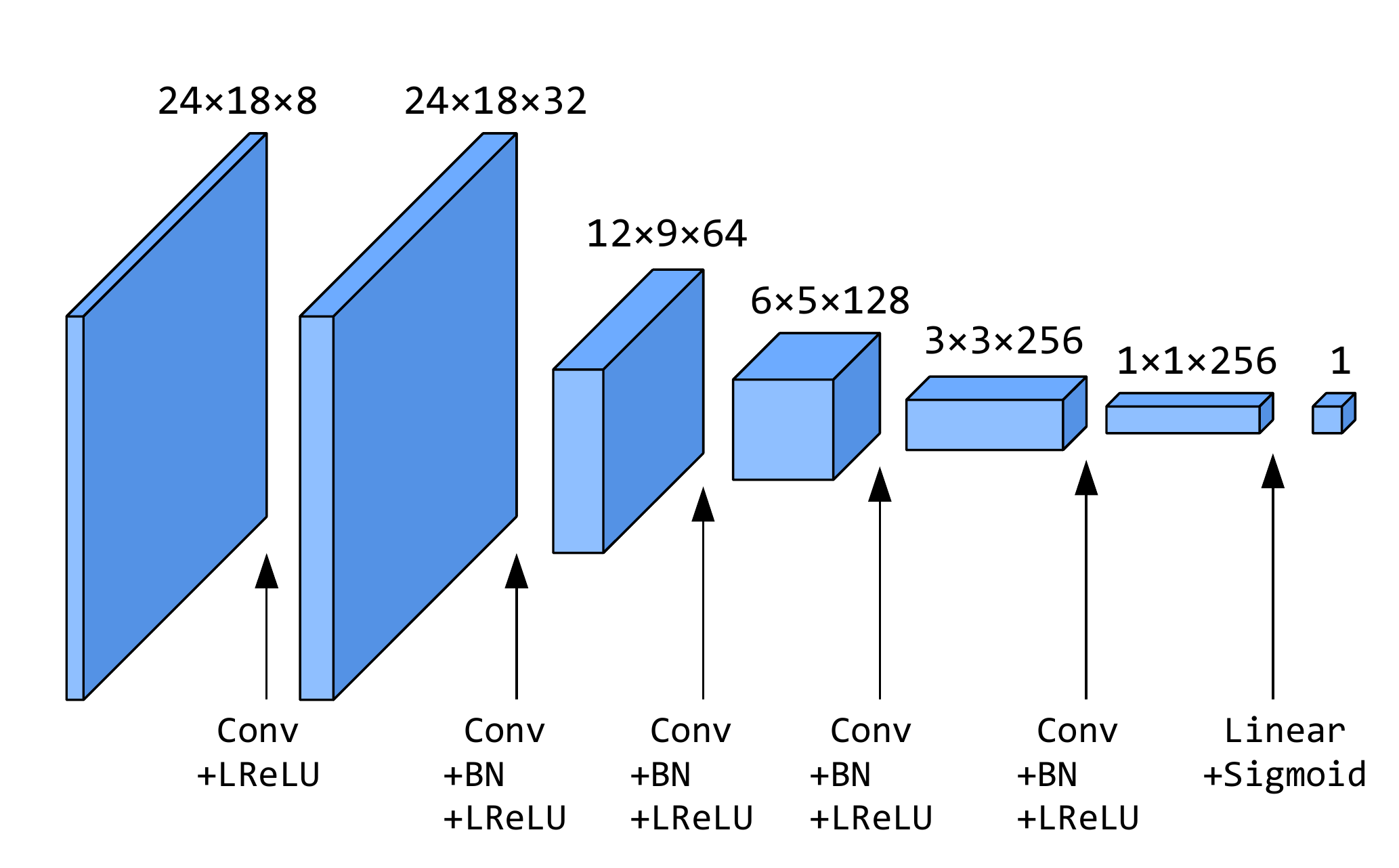}%
    \label{figure_architecture_discriminator}}
    
    \caption{Detailed description of our network architecture. Notations:
    Convolution (Conv), Transposed Convolution (Tr.Conv), Batch Normalization (BN), Rectified Linear Unit (ReLU) and Leaky Rectified Linear Unit (LReLU).}
    \label{figure_architecture}
\end{figure*}


    
    \section{Related Work} \label{section_related_work}
        \subsubsection*{Deep generative inpainting}
    Deep generative models have demonstrated impressive performance on image inpainting \cite{paper_inpainting, paper_cvaegan, paper_dual_discriminator_inpainting, paper_self_attention_inpainting}. For this task, the need to faithfully reproduce the visible surroundings of missing image regions adds an additional constraint to the generative synthesis process. Yeh \emph{et al.} \cite{paper_inpainting} rely on a \GAN{} \cite{paper_gan} whose generator is fed with noise samples from a known prior distribution. In the inference stage, the closest noise sample to the corrupted image is obtained by backpropagating the gradients of the generator network and fed to the generator to obtain the inpainted image. In their seminal paper \cite{paper_seminal_GAN_inpainting}, Pathak \emph{et al.} take a different approach that combines a \GAN{} and a context encoder. The latter network is trained to reconstruct a full image from an input with missing parts. It acts as the \GAN{} generator, while the discriminator enforces the plausibility of the inpainting result. Iizuka \emph{et al.} \cite{paper_dual_discriminator_inpainting} enrich this architecture with two discriminators that separately process small-scale and large-scale image texture. Yu \emph{et al.} \cite{paper_self_attention_inpainting} further add a self-attention module to better take advantage of distant image patches to fill the missing regions. Bao \emph{et al.} \cite{paper_cvaegan} replace the context encoder with a \VAE{} \cite{paper_vae} and incorporate an image classifier to specialize the generative process to sub-categories.
    
\subsubsection*{Human motion inpainting and in-betweening}
    Past research on human pose and motion inpainting has mostly focused on in-betweening, \emph{i.e.}, the process of temporally interpolating character poses between input keyframes manually edited by an artist. Harvey \emph{et al.} \cite{paper_inbetweening_Harvey} and Zhou \emph{et al.} \cite{paper_inbetweening_Zhou} propose deep learning schemes to perform this task. The inputs consist of 3D character poses where all joint positions are known, and the inpainting is performed only along the temporal dimension.  The deep networks are extensions of an \ERD{} model originally proposed in \cite{paper_ERD}, in which the temporal motion context is captured by a Recurrent Neural Network that operates in the embedding at the output of the decoder. More closely related to our work is the approach of Hernandez Ruiz \emph{et al.} \cite{paper_motion_inpainting}. They complement the \ERD{} architecture with a \GAN{} whose generator operates in the latent space of the encoder. Their scheme can both temporally interpolate motion frames and recover missing joints within a pose.\par

    Unlike all these approaches that are fed with 3D poses, our scheme takes as input 2D joint locations that could typically be obtained from a 2D human pose estimation framework. We believe 2D pose inpainting is more challenging: although depth information is missing at the input, the depth dimension needs to be accounted for by the inpainting process in order to obtain 2D joint location estimates that are consistent with the motion of the character in 3D space.

    \section{Method} \label{section_method}
        \begin{figure*}[t]
    \centering
    
    \subfloat[Flow during discriminator's training step.]{\includegraphics[width=2.6in]{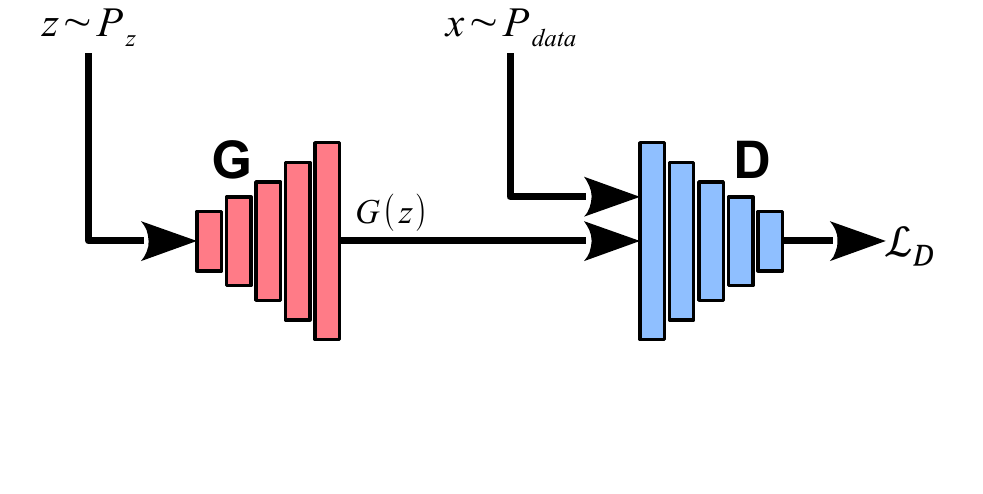}%
    \label{figure_flows_dis}}
    \hfil
    \subfloat[Flow during generator/encoder's training step.]{\includegraphics[width=4.42in]{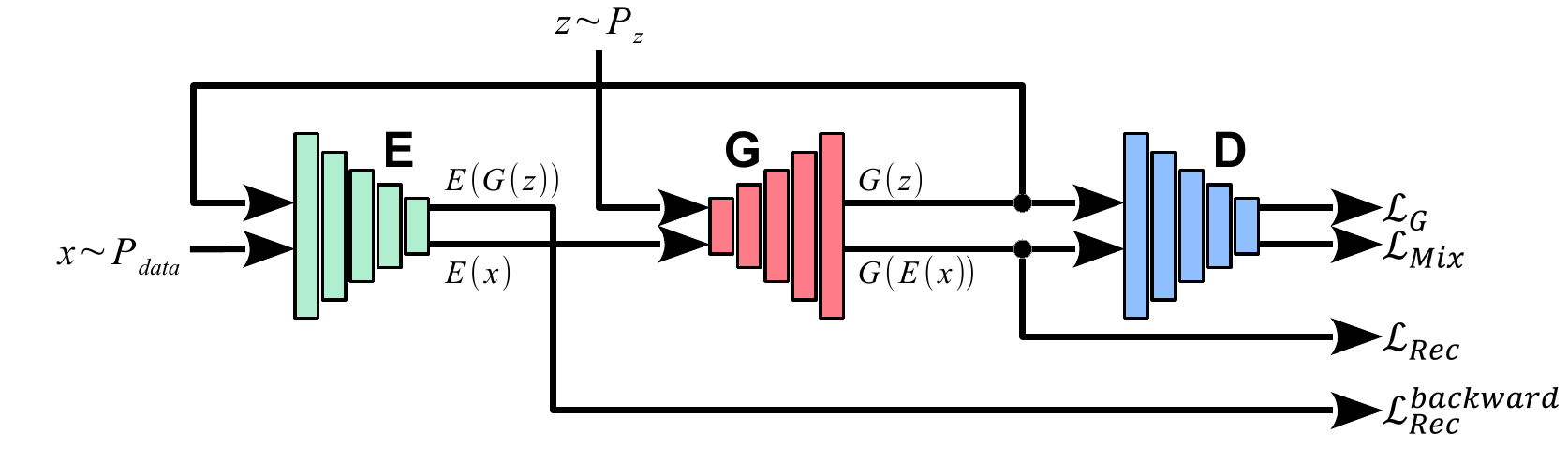}%
    \label{figure_flows_ae}}
    
    \caption{Flows through our combined AE-GAN model during training.}
    \label{figure_flows}
\end{figure*}
        \subsection{Overview}
    We propose a method to upsample and inpaint an animated skeleton to infer the locations of missing or unseen joints and provide a higher-resolution representation of the body pose. To this purpose we leverage a deep generative network that we train with moving skeleton sequences, rather than static poses, in order to better disambiguate the estimation of missing joint locations.\\
    As illustrated in Figure \ref{figure_network}, our model consists of a \GAN{} coupled with an encoder, both forming an autoencoder where the generator plays the role of the decoder. It benefits from the generative power of \GAN{}s and mitigates instability during training by introducing supervision from the encoder.

\subsection{Detailed Architecture}  
    Our network conforms to the architecture of \DCGANs{} \cite{paper_dcgan}, using fractionally-strided transposed convolutions in the generator and strided convolutions in the discriminator (see Figure \ref{figure_architecture}). \DCGANs{} also use Rectified Linear Units (ReLU) as activation functions in the generator and Leaky ReLU (LReLU) in the discriminator. Moreover, Batch Normalization (BN) is also applied after almost each convolutional layer. Except for the output size of its final layer, the encoder has the same architecture as the discriminator.

\subsection{Training}
    In this section we describe the representation used for joint position data, the loss functions and the optimization procedure.
    \subsubsection*{Data Representation}
        A $2D$ pose sequence is usually represented as a $3$-dimensional tensor containing the $2D$ coordinates of each joint at each frame. To obtain meaningful and efficient convolutions, we rearrange the joints as shown in Figure \ref{figure_data_representation}.
        In this representation, each entry holds the $2D$ coordinates for two joints (\emph{i.e.}, four channels). Symmetric joints (\emph{e.g.}, feet, knees, \emph{etc.}) are paired to form an entry, while joints in the axial skeleton (\emph{e.g.}, pelvis, thorax, \emph{etc.}) are duplicated\footnote{Both discriminator and encoder duplicate axial input joints while the generator produces duplicated axial joints in a first step and then outputs the average of the two versions.} in order to obtain consistent four-channel entries. This reformatting of data to a rectangular $2D$ grid allows to use regular $2D$ convolutions in our deep network.
    
    \subsubsection*{Notation}
        In the following, we note $E$, $G$ and $D$ the encoder, the generator and the discriminator networks, respectively. In addition $P_{z}$ and $P_{data}$ denote the latent and the data distributions respectively. Finally, $P_{\tilde{x}}$ stands for the distribution of uniformly sampled points along straight lines between pairs of points sampled from the data distribution $P_{data}$ and the generator distribution, \emph{i.e.} $P_{z}$ mapped through $G$, as defined in \cite{paper_improved_wgan}.\par
    
    \subsubsection*{Adversarial Loss}        
        Traditionally, a \GAN{} consists of a generator and a discriminator. The former is trained to produce realistic samples while the latter aims at distinguishing those from real samples, both competing against each other. The ability to generate realistic samples can be expressed more formally as the similarity between two probability distributions that are the data distribution and the distribution of samples produced by the generator. The original formulation of \GANs{} \cite{paper_gan} measures the similarity with the Jensen–Shannon divergence. However, this divergence fails to provide meaningful values when the overlap between the two distributions is not significant which often makes \GANs{} quickly diverge during training. Arjovsky et al. \cite{paper_wgan} introduced \WGANs{}, showing that, under the hypothesis that the discriminator is 1-Lipschitz, the Jensen–Shannon divergence can be replaced by the Wasserstein distance that have better properties for convergence. Then, Gulrajani et al. \cite{paper_improved_wgan} propose a gradient penalty term in the \WGAN{} loss function to enforce the 1-Lipschitz hypothesis on the discriminator.\par
        Therefore, we opt for the gradient-penalized \WGAN{} and have the following loss functions for the generator and the discriminator, respectively:
        \begin{IEEEeqnarray}{rCl}
            \label{equation_generator_loss}
            \mathcal{L}_{G} = && \expectation[2]{z \sim P_{z}}{-D(G(z))}\\
            \mathcal{L}_{D} = && \expectation[2]{z \sim P_{z}}{D(G(z))}
            - \expectation[2]{x \sim P_{data}}{D(x)}
            \nonumber\\
            && +\> \lambda_{gp} ~ \expectation[2]{\tilde{x} \sim P_{\tilde{x}}}{
                \parentheses[2]{\norm[1]{\nabla_{\tilde{x}} D(\tilde{x})}_2 - 1}^2
            }
        \end{IEEEeqnarray}
        where $\lambda_{gp}$ is the gradient penalty coefficient.\pagebreak
        
        \subsubsection*{Reconstruction Losses}
        Like autoencoders, our model is encouraged to reconstruct inputs that are encoded and then decoded through a reconstruction loss minimizing differences between inputs and outputs. We also incite our model to be consistent when generating and then encoding from latent codes sampled from the prior distribution with a backward reconstruction loss, as in cycle-consistent \VAEs{} \cite{paper_cyclic_vae}. Such backward reconstruction loss facilitates the convergence but more importantly enforces the distribution of the encoder outputs to match the prior distribution $P_{z}$ imposed on our \GAN{}. As illustrated in Figure \ref{figure_flows_ae}, the total loss in the autoencoding scheme is
        \begin{IEEEeqnarray}{rCl}
            \mathcal{L}_{AE} = \mathcal{L}_{Rec} + \mathcal{L}_{Rec}^{backward}
        \end{IEEEeqnarray}
        \par
		
		$\mathcal{L}_{Rec}$ is itself made up of two terms penalizing respectively the joints position and velocity errors of the reconstructed sample $\hat{x} = G(E(x))$ with respect to the ground truth $x$. More formally, we use the \MPJPE{} \cite{paper_human36m} to quantify joint position errors:
        \begin{IEEEeqnarray}{c}\label{equation_mpjpe}
            MPJPE(x, \hat{x}) = \frac{1}{J} \frac{1}{F} \sum_{f, j} \norm[1]{(x - \hat{x})_{j,f}}_2
        \end{IEEEeqnarray}
        where $j$ and $f$ denote the joint and frame considered; $J$ and $F$ are the numbers of joints and frames respectively.\\
		In analogy to the \MPJPE{}, we define the \MPJVE{} as
        \begin{IEEEeqnarray}{c}\label{equation_mpjve}
            MPJVE(x, \hat{x}) =
            \frac{1}{J} \frac{1}{F} \sum_{f, j} \norm[2]{\parentheses[1]{v(x) - v(\hat{x})}_{j, f}}_2
        \end{IEEEeqnarray}
        where $v(\cdot)$ computes the velocity of each joint at each frame as the position difference between the current and previous frame. This secondary term penalizing velocity errors acts as a powerful regularizer that accelerates the convergence in early iterations and also reduces temporal jitter in the joint locations of the generated pose sequences. Hence, $\mathcal{L}_{Rec}$ is the weighted sum of Eq. (\ref{equation_mpjpe}) and Eq. (\ref{equation_mpjve}): 
        \begin{IEEEeqnarray}{rCl}
            \mathcal{L}_{Rec} = 
            & & \lambda_{p} ~ \expectation[2]{x \sim P_{data}}{
                MPJPE(x, \hat{x})
            } \nonumber \\
            & +\> & \lambda_{s} ~ \expectation[2]{x \sim P_{data}}{
                MPJVE(x, \hat{x})
            }
        \end{IEEEeqnarray}
        where $\lambda_{p}$ and $\lambda_{s}$ are the weights.\par\vfill\null
        
		The second component $\mathcal{L}_{Rec}^{backward}$ of our autoencoder's objective focuses on the reconstruction of the latent code $z$ sampled from the prior distribution $P_z$. It minimizes the \MSE{} between $z$ and its reconstructed version $\hat{z} = E(G(z))$:
        \begin{IEEEeqnarray}{rCl}
            \mathcal{L}_{Rec}^{backward} =
            \lambda_{z} \expectation[2]{
                z \sim P_{z}}{MSE(z, \hat{z})
            }
        \end{IEEEeqnarray}
        \subsubsection*{Mixed Loss}
        We further encourage the generation of realistic sequences by adding a loss term to penalize unrealistic reconstructed pose sequences. Here we make use of the discriminator to tell both the generator and the encoder whether the reconstructed pose sequence $\hat{x} = G(E(x))$ is realistic or not. We use the same formulation as for the generator adversarial loss (see in Eq. \ref{equation_generator_loss}) but applied to $\hat{x}$ instead of $G(z)$:
		\begin{IEEEeqnarray}{rCl}
            \mathcal{L}_{Mix} = \lambda_{m} ~ \expectation[2]{x \sim P_{data}}{- D(\hat{x})}
        \end{IEEEeqnarray}
        
    \subsubsection*{Optimization}
         In summary, the encoder, the generator and the discriminator are optimized w.r.t. the loss functions $\mathcal{L}_{AE} + \mathcal{L}_{Mix}$, $\mathcal{L}_{G} + \mathcal{L}_{AE} + \mathcal{L}_{Mix}$ and $\mathcal{L}_{D}$, respectively. Similarly to a \GAN{}, during the training we optimize at each iteration the discriminator in a first step and then the generator and the encoder. Figure \ref{figure_flows} illustrates the computational flows through the network during both training steps.

	\subsubsection*{Temporal Variance Regularization}
		\GAN{}s are known to produce sharp samples, but for the considered task this can lead to perceptually disturbing temporal jitter in the output pose sequences. To optimize the tradeoff between sharpness and temporal consistency, we feed the discriminator with stacked joint positions and velocities (computed for each joint at each frame as the position difference between the current and previous frame). The velocities favour the rejection of generated samples that are either temporally too smooth or too sharp. This idea is conceptually inspired from \cite{paper_pggan}, where the variation of generated samples is increased by concatenating minibatch standard deviations at some point of the discriminator.

\begin{table}[!t]
    \renewcommand{\arraystretch}{1.0}
    \caption{
        Results of the joint upsampling experiments. We upsample back to 28 joints a ground truth 2D pose sequence purposedfully downsampled to 12 joints. Removing Procrustes alignment (w/o P.A.) and the encoder (w/o ENC.) substantially degrades performance.
        See the text for the definition of the performance metrics.
    }
    \label{table_upsampling}
    
    \centering

    \begin{tabular}{| l | c c c c |}
        \hline
        method & \scriptsize PCKh@0.1 & \scriptsize PCKh@0.5 & \scriptsize PCKh@1.0 & \scriptsize AUC \\
        \hline
        \hline
        JUMPS w/o P.A. & 0.0368 & 0.4384 & 0.6814 & 0.3912 \\
        JUMPS w/o ENC. & 0.1701 & 0.8259 & 0.9678 & 0.7005 \\
        JUMPS w/o overlap & 0.5821 & 0.9648 & 0.9962 & 0.8727 \\
        JUMPS & \textbf{0.6096} & \textbf{0.9674} & \textbf{0.9965} & \textbf{0.8803} \\
        \hline
	\end{tabular}
\end{table}

\pagebreak

\subsection{Inference} \label{section_inference}
    We leverage the human motion model learnt by the generator to recover missing joints in an input pose sequence $x$. Given $x$, we optimize $z$ using gradient backpropagation across the generator network
    of a contextual loss that minimizes the discrepancy between $G(z)$ and $x$ on available joints. To this contextual loss we add a prior term that maximizes the discriminator score on the generated pose sequence. This process is closely related to the semantic image inpainting approach in \cite{paper_inpainting}; however we take advantage of our encoder to compute a starting latent code $z = E(x)$ as in \cite{paper_cvaegan}. This approach also applies to upsampling, considering that the added joints are missing in the input.\\
    Formally, we first solve 
    \begin{IEEEeqnarray}{c}
        z^* = \argmin_z{\mathcal{L}_{Inp}(x, z)}
    \end{IEEEeqnarray}
    by gradient descent where $\mathcal{L}_{Inp}$ is our inpainting objective function composed of a contextual loss and a prior loss. Then, we generate $x^*=G(z^*)$ that best reconstructs $x$ w.r.t $\mathcal{L}_{Inp}$.
    
    \subsubsection*{Inpainting Loss Function}
        Our contextual loss minimizes the weighted sum of \MPJPE{} and \MPJVE{} between the input pose sequence $x$ and the generated pose sequence $\hat{x} = G(z)$. Additionally, the prior loss maximizes the discriminator score $D(\hat{x})$ on the generated pose sequence:
        \begin{IEEEeqnarray}{rCl}
            \mathcal{L}_{Inp} = ~ && \underbrace{\gamma_{p} ~ MPJPE(x, \hat{x}) + \gamma_{s} ~ MPJVE(x, \hat{x})}_\text{contextual loss} \nonumber\\
            && \underbrace{-\> \gamma_{d} ~ D(\hat{x})}_\text{prior loss}
        \end{IEEEeqnarray}

    \subsubsection*{Post-Processing}
        At each gradient descent step, we generate the pose sequence $\hat{x} = G(z)$. At this point, we additionally use the fact that we are given a pose sequence $x$ to be inpainted by optimally translating, scaling and rotating $\hat{x}$ to match $x$. This process (known as Ordinary Procrustes Analysis)
        has a low overhead but makes the gradient descent convergence several times faster and improves inpainting results.
        
    \subsubsection*{Pose Sequence Length}
        Our deep network requires pose sequences to have a constant number of frames $F$. Here we describe a simple mechanism to handle longer variable-length pose sequences. Given a pose sequence $x$ longer than $F$ frames, the idea is to independently inpaint fixed-length subsequences of $x$ and then concatenate the results into a single inpainted pose sequence having the same length as $x$.
        Using this process there is no guarantee that two consecutive subsequences will be smoothly concatenated. To prevent such discontinuities in the generated sequences we use half overlapping subsequences. At each temporal sample where an overlap is present we select among the candidate inpainted frames the one closest to the input, in the sense of the minimal contextual loss term in $\mathcal{L}_{Inp}$.

    \begin{table}[!t]
    \renewcommand{\arraystretch}{1.0}
    \caption{
        Results of the Alpha Pose post-processing experiments. We perform inpainting and upsampling to 28 joints of 2D pose estimates obtained by running Alpha Pose on video sequences. The ablation studies confirm the conclusions drawn from the joint upsampling experiments (see table \ref{table_upsampling}).
    }
    \label{table_hpe}
    \centering

    \begin{tabular}{| l | c c c c |}
        \hline
        method & \scriptsize PCKh@0.1 & \scriptsize PCKh@0.5 & \scriptsize PCKh@1.0 & \scriptsize AUC \\
        \hline
        \hline
        \emph{AlphaPose} & \textbf{0.0941} & 0.7659 & 0.9157 & 0.6310 \\
        JUMPS w/o P.A. & 0.0207 & 0.3423 &  0.6304 & 0.3249 \\
        JUMPS w/o ENC. & 0.0537 & 0.6801 & 0.9059 & 0.5692 \\
        
        JUMPS w/o overlap & 0.0831 & 0.7701 & \textbf{0.9277} & 0.6326 \\
        JUMPS & 0.0842 & \textbf{0.7723} & 0.9276 & \textbf{0.6341} \\
        \hline
	\end{tabular}
\end{table}

    \section{Experiments} \label{section_experiments}
        \begin{figure*}[!t]
    \centering
    
    \subfloat[
        JUMPS w/o overlap
    ]{
        \includegraphics[width=\columnwidth]{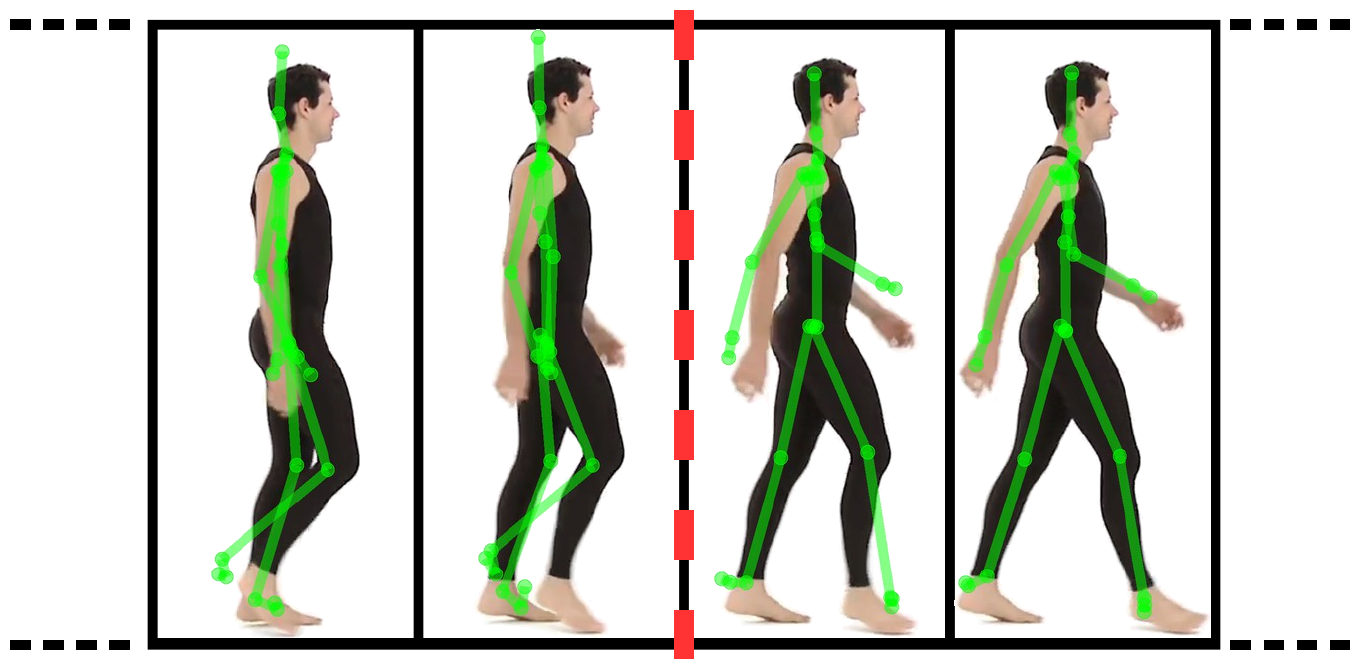}%
    }
    \hfil
    \subfloat[
        JUMPS
    ]{
        \includegraphics[width=\columnwidth]{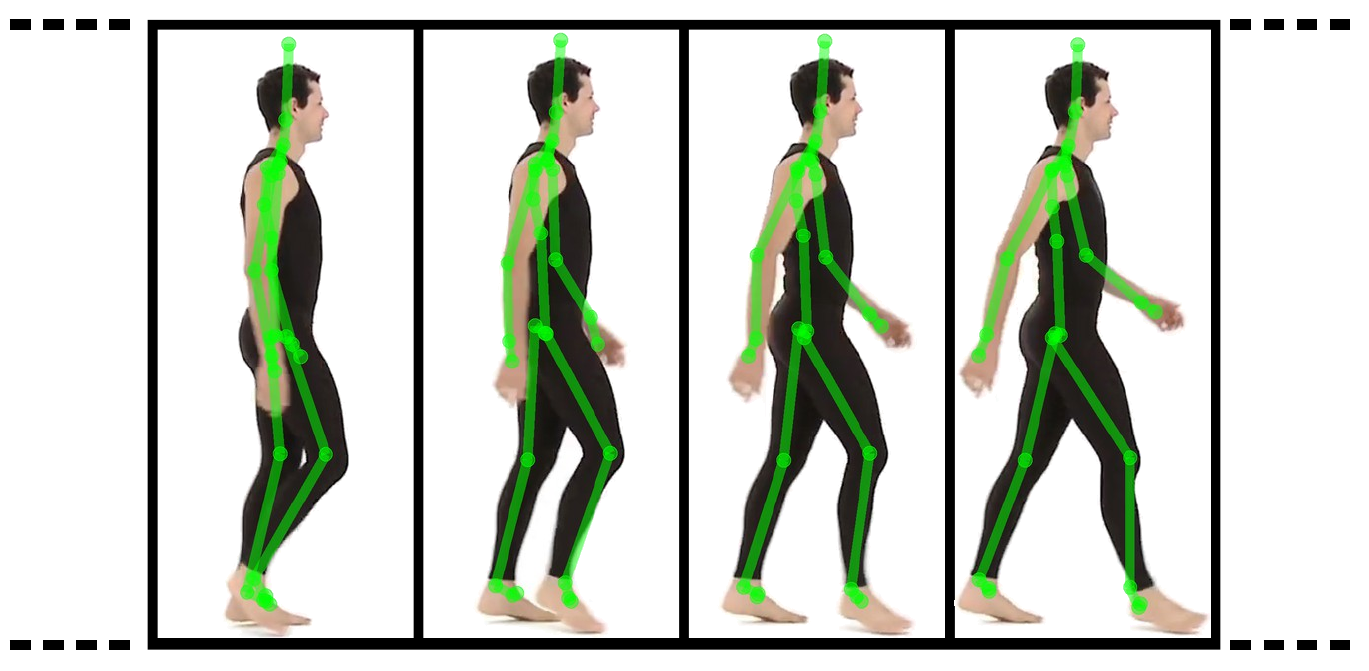}%
    }
    
    \caption{
        Qualitative example of the influence of overlapping temporal chunks. This subset of four consecutive frames in a longer sequence is inpainted with no overlap (left) and half overlap (right). The frames are located at the end (first two) and the beginning (last two) of two consecutive chunks. Note the temporal discontinuity of joint locations (head top, forearms, feet) in the inpainted sequence at the chunk boundary with no overlap (dashed red line, left). The temporal consistency is much better using an overlap (right).
    }
    \label{figure_overlap}
\end{figure*}

\subsection{Datasets and Metrics}
	\subsubsection*{Training and test sets}
        We rely on \emph{MPI-INF-3DHP} \cite{paper_mpi_inf_3dhp} for our experiments. This dataset contains image sequences in which $8$ actors perform various activities with different sets of clothing. This dataset is well suited for our task of joints upsampling since it is one of the public databases having the highest skeleton resolution, i.e. skeletons with 28 joints. Since our method focuses on fixed length $2D$ pose sequences, we generated a set of around $835K$ $2D$ pose sequences of $24$ frames (\emph{i.e.}, $F$=$24$) each using projections of the original $3D$ pose data from randomized camera viewpoints. We also selected around $166K$ images annotated with $2D$ poses directly from \emph{MPI-INF-3DHP} with no preprocessing for testing.
        
    \subsubsection*{Evaluation Metrics}
        We report our experiments results with the \PCKh{} \cite{paper_mpii} and the \AUC{} \cite{paper_auc} metrics. \PCKh{} metric consider a joint as correct if its distance to the ground truth normalized by head size is less than a fixed threshold and the \AUC{} aggregates \PCKh{} over an entire range of thresholds. We use the common notation \PCKh{}@$\alpha$ to refer to \PCKh{} with threshold $\alpha$ and we compute the \AUC{} over the range $[0, 1]$ of thresholds.

\subsection{Implementation Details}
    Our deep network (see Figure \ref{figure_architecture} for detailed architecture) has about $3$ millions learnable parameters that are almost equally distributed over the encoder ($1'148'096$), the generator ($1'148'480$) and the discriminator ($1'115'393$). Our implementation is in Python and deeply relies on PyTorch library. Training and experiments have been executed on a NVIDIA Tesla P100 PCIe 16GB.

	\subsubsection*{Training}
	    We trained our model for $60$ epochs (about 11 hours) with a minibatch size of $256$ using the Adam algorithm \cite{paper_adam} with optimization hyperparameters $\alpha = 0.0001$, $\beta_{1} = 0.5$, $\beta_{2} = 0.999$ and $\epsilon = 10^{-8}$. We followed the suggestions for \DCGANs{} from \cite{paper_dcgan} to reduce (w.r.t. \cite{paper_adam} suggestions) $\alpha$ and $\beta_{1}$. 
	    As in \cite{paper_dcgan}, we observed that $\beta_{1} = 0.5$ helped to stabilize the training.\par
	    
	    We set the Wasserstein gradient penalty weight $\lambda_{gp}$ to $10$ as proposed in \cite{paper_improved_wgan}, and our loss weights $\lambda_p$, $\lambda_s$, $\lambda_z$ and $\lambda_m$ to $200$, $100$, $2$ and $1$ respectively. We empirically found these values to work well.

	\subsubsection*{Inference}
		We compute the latent code again using the Adam optimization algorithm with $\alpha = 1$, $\beta_1 = 0.8$, $\beta_2 = 0.999$ and $\epsilon = 10^{-8}$. The weights of the inpainting loss are set to $\gamma_p = 10$, $\gamma_s = 5$ and $\gamma_d = 15$. We stop the optimization after 200 iterations. These hyperparameter values has been chosen to make the optimization in a limited number of iterations and avoid matching noise or imperfections in inputs.\par
	    
	    To improve inference results we perform several optimizations of the latent code in parallel for a single input, starting from different initializations. One of these starting points is computed as the output of the encoder fed by the input pose sequence, the others are randomly sampled from the prior distribution. We keep the one closest to the input, in the sense of the inpainting loss $\mathcal{L}_{Inp}$.

\subsection{Joints Upsampling}
    Our first experiment focuses on the upsampling task. We downsample ground truth $28$-joint pose sequences to $12$ joints that are common to the \emph{MPI-INF-3DHP} dataset and \emph{AlphaPose} skeletons (see Figure \ref{figure_occlusion} left), upsample them back to $28$ joints using our method, and compare the result to the original sequence. Table \ref{table_upsampling} provides \PCKh{} and \AUC{} values for this experiment. Assuming a typical human head size, the positioning error is less than $2.25$ \emph{cm} for half of the upsampled joints (\PCKh{} threshold = $0.1$) and less than $11.25$ \emph{cm} for $95\%$ of them (\PCKh{} threshold = $0.5$).

\subsection{2D Human Pose Estimation}
\begin{figure*}[!t]
	\centering
	\includegraphics[width=\columnwidth]{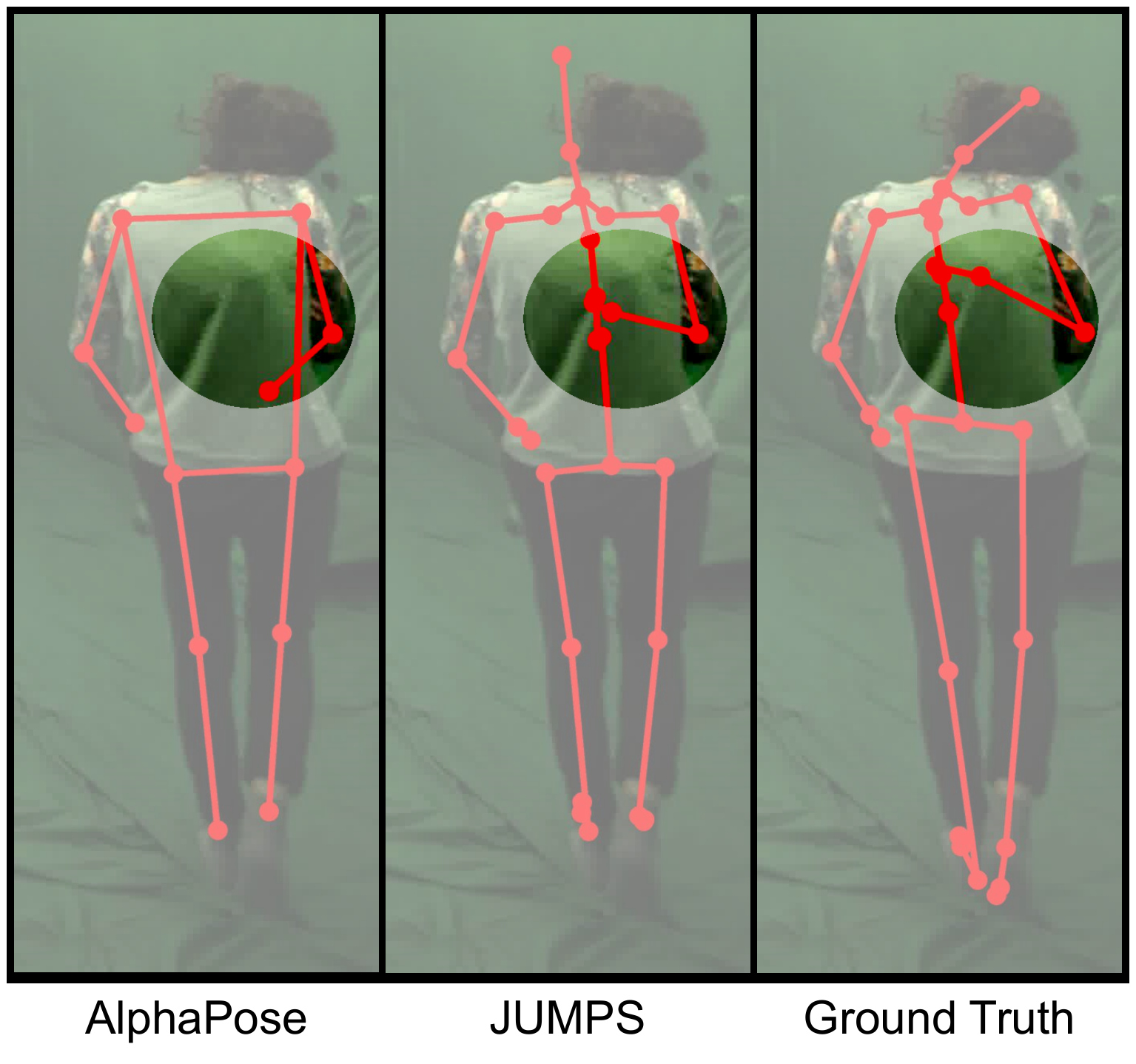}			
	\caption{Example of a limb (right forearm) occluded by subject's body inaccurately estimated by \emph{AlphaPose} but recovered by our method based on human motion priors. Note that these images have been intentionally whitened except for the area around the occlusion for clarity purposes.} 
	\label{figure_occlusion}
\end{figure*}

    Our second experiment deals with the concrete use case of inpainting and upsampling joints on a pose sequence obtained using 2D Human Pose Estimation. We rely on \emph{AlphaPose}\footnote{Implementation based on \cite{paper_rmpe, paper_crowdpose, paper_pose_flow} available at \emph{https://github.com/MVIG-SJTU/AlphaPose}} to preprocess videos in our test set. \emph{AlphaPose} provides $12$-joint pose estimates that we post-process using our method to recover missing (\emph{e.g.}, occluded) joints and upsample to $28$ joints. Table \ref{table_hpe} summarizes the results for this experiment. The positioning accuracy is roughly the same for the inpainted / upsampled joints and for the joints obtained by Human Pose Estimation. Thus, our method enriches the pose information without sacrificing accuracy. Figure \ref{figure_occlusion} illustrates how our method is able to correct the right wrist position mispredicted by \emph{AlphaPose} based on the temporal consistency of the right forearm movement.

\subsection{Ablation Studies}

    \subsubsection{Procrustes Analysis}
        line \emph{JUMPS w/o P.A.} in tables \ref{table_upsampling} and \ref{table_hpe} gives the joint positioning accuracy when the Procrustes Analysis post-processing of our method (see section \ref{section_inference}) is removed. Instead we map all pose sequences to the image frame using the same affine transform. Rigidly aligning the generated poses during the gradient descent optimization of the latent code is critical to the performance of our approach.

    \subsubsection{Encoder}
        as shown by the accuracy estimates in lines \emph{JUMPS w/o ENC.} of the same tables, removing the encoder in front of the \GAN{} in our architecture, during both training and inference stages, substantially degrades performance. The encoder regularizes the generative process and improves the initialization of the latent code at inference time, yielding poses that better match the available part of the input skeleton.

    \subsubsection{Overlapping subsequences}
        Processing input sequences with an overlap yields only a slight improvement of performance over no overlap, the gain being stronger at high accuracy levels. Indeed, since the optimization of the latent code in our method matches the upsampled pose to the input, an additional selection of the result closest to the input among the several candidate poses at each frame when using an overlap brings little gain in accuracy.

        However, as illustrated on Figure \ref{figure_overlap}, we found that processing overlapping chunks of frames noticeably improves the temporal consistency of the output pose sequence. We observed that the per-frame joint positioning accuracy drops at the extremities of the processed chunks, probably because of the reduced temporal context information there. Without overlap this introduces an increased temporal jitter at the chunk boundaries of the generated pose sequence, which is likely to incur perceptually disturbing artifacts when applying our method to, \emph{e.g.}, character animation.

    \section{Conclusion} \label{section_conclusion}
        In this paper we presented a novel method for 2D human pose inpainting focused on joints upsampling. Our approach relies on a hybrid adversarial generative model to improve the resolution of the skeletons (\emph{i.e.}, the number of joints) with no loss of accuracy. To the best of our knowledge, this is the first attempt to solve this problem in 2D with a machine learning technique. We have also shown its applicability and effectiveness to Human Pose Estimation.\\
Our framework considers a $12$-joint $2D$ pose sequence as input and produces a valuable $28$-joint $2D$ pose sequence by inpainting the input. The proposed model consists of the fusion of a deep convolutional generative adversarial network and an autoencoder. Ablation studies have shown the strong benefit of the autoencoder, since it provides some supervision that greatly helps the convergence and accuracy of the combined model. Given an input sequence, inpainting is performed by optimizing the latent representation that best reconstructs the low-resolution input. The encoder provides the initialization and a prior loss based on the discriminator is used to improve the plausibility of the generated output.\\
The obtained results are encouraging and open up future research opportunities. Better consistency of the inpainted pose sequences with true human motion could be obtained either by explicitly enforcing biomechanical constraints, or by extending the method to $3D$ joints, in order to benefit from richer positional information on the joints. Additionally, a potentially fruitful line of research would be to tackle as a whole, from a monocular image input, the extraction of $3D$ human pose and the upsampling of skeleton joints. Finally, we plan to study more genuine temporal analysis by using a different network architecture handling either longer or variable-length pose sequences (\emph{e.g.}, based on recurrent neural networks or fully convolutional networks).
    

\end{document}